\documentclass[conference]{IEEEtran}
\IEEEoverridecommandlockouts
\usepackage{cite}
\usepackage{amsmath,amssymb,amsfonts}
\usepackage{algorithm}
\usepackage{algpseudocode}
\usepackage{graphicx}
\usepackage{textcomp}
\usepackage{xcolor}
\usepackage{balance}

\usepackage[utf8]{inputenc}
\usepackage{array}
\usepackage{wrapfig}
\usepackage{multirow}

\def\BibTeX{{\rm B\kern-.05em{\sc i\kern-.025em b}\kern-.08em
		T\kern-.1667em\lower.7ex\hbox{E}\kern-.125emX}}
\begin{document}
	\title{Vision-Based Incoming Traffic Estimator Using Deep Neural Network on General Purpose Embedded Hardware}
	\author{\IEEEauthorblockN{K.~G.~Zoysa, and S.~R.~Munasinghe~\IEEEmembership{SMIEEE}}
		\IEEEauthorblockA{\textit{Dept. of Electronic and Telecommunication Engineering} \\
			\textit{University of Moratuwa}\\
			Moratuwa 10400, Sri Lanka \\
			rohan@uom.lk}
		\thanks{S. R. Munasinghe is a visiting Fellow at the Dept of Global Development, College of Agriculture and Life Sciences, Cornell University, NY, USA}}
	\maketitle
	\begin{abstract}
		Traffic management is a serious problem in many cities around the world. Even the suburban areas are now experiencing traffic congestion regularly. Inappropriate traffic control leads to waste of fuel, time, and productivity of nations. Though the traffic signals are used to improve traffic flow, they often cause problems due to inappropriate or obsolete timing that does not tally with the actual traffic intensity at the intersection. Traffic intensity determination based on statistical methods only gives the average intensities expected at any given time. However, to control traffic accurately, it is required to know the real-time traffic intensity. In this research, image processing and machine learning have been used to estimate actual traffic intensity in real time. General-purpose electronic hardware has been used for in-situ image processing based on the edge-detection method. A deep neural network (DNN) was trained to infer traffic intensity in each image in real time. The trained DNN estimated traffic intensity accurately in 90\% of the real-time images during road tests. The electronic system was implemented on a Raspberry Pi single-board computer, hence it is cost-effective for large-scale deployment. 
	\end{abstract}
	
	\section{Introduction}
	In most parts of the world, particularly in urban cities vehicular traffic increases steadily. Traffic intensity fluctuation during the day and over a week is also difficult to estimate. Traffic management and control, despite the advancement of road technologies, has not been able to cope with the increasing complexity of road traffic. This has led to increasingly inappropriate traffic control. The situation is aggravating in almost all urban cities, particularly in the developing world where it is extremely difficult to predict the variations in traffic intensity. The inappropriate traffic control results in slow vehicle speed, fuel wastage, time waste, and air pollution as well. Some countries have already reported issues such as urban air pollution, health issues, and long-term effects on the economy and development. In this view, accurate traffic control has become a top priority for those countries \cite{TrafficCongestionEconomicImpacts}. During the last decade, different technologies including vision-based methods have been introduced to manage traffic \cite{ASurveyofVisionBasedTrafficMonitoringofRoadIntersections}. Vision-based systems could be very effective because they could provide real-time traffic information which is key to effective traffic control. The major reason for inappropriate traffic control at intersections is the preprogrammed static timing which opens and closes the lanes without knowing the actual traffic situation at that moment.\par
In the case of traffic control using static time schedule\cite{Static_traffic_assignment_with_queuing} the day is split into several time slots and in each slot, the average traffic intensity is measured from which the traffic signal times for that slot are calculated. This method is widely practiced though it is often found incorrect due to inherent statistical variations. There have been several traffic detection methods including loop detectors \cite{Loop_detector_for_traffic_signal_control} \cite{A_vehicle_classification_based_on_inductive_loop_detectors}, use of GPS data to determine queue length and control delays \cite{AnalysisofIntersectionQueueLengthsandLevelofServiceUsing_GPS_data}, use of sensor networks \cite{Sensor_networks_for_monitoring_traffic} to detect vehicles, and use of reference images \cite{Density_based_smart_traffic_control_system_using_canny_edge_detection_algorithm_for_congregating_traffic_information} to estimate traffic intensity at the cost of increased computational overhead and delay in traffic sensing. In \cite{Videobasedsystemdevelopmentforautomatictrafficmonitoring}, density and speed of traffic are estimated to classify traffic into only three categories; free flow, slow moving, and congestion. In \cite{ObjectCountingonLowQualityImages} and \cite{A_YOLOBased_Traffic_Counting_System}, vehicles are counted using you-only-look-once (YOLO) detector. The accuracy of the YOLO detector is very high, however, it is computationally intensive and perhaps not cost-effective for actual traffic control on a large scale. In \cite{A_reliable_counting_vehicles_method_in_traffic_flow_monitoring}, a robust algorithm has been introduced to count vehicles.\par
In this research, the objective is to devise a cost-effective yet accurate method to estimate real-time traffic intensity. Therefore, a Raspberry Pi single-board computer, which is a general-purpose electronic hardware that can run low-complexity machine learning algorithms was used for image processing and traffic intensity estimation. The reputed edge-detection method and supervised learning method were used for traffic intensity estimation in a deep neural network. A set of annotated traffic images was used for the DNN training. The major concern was the accuracy of the traffic intensity estimate and the time taken to derive that result while keeping the device cost-effective for actual implementation on a large scale.
	\section{Vision-Based Real-time Traffic Intensity Determination}\label{AA}
	One of the major requirements in using vision-based methods for traffic intensity determination is to make it fast enough for real-time implementation. On the other hand, sophisticated vision-based methods, though very accurate and fast, are financially not viable for large-scale deployment. Therefore, it is required to develop a new vision-based traffic intensity estimation algorithm that can run on general-purpose electronic hardware while meeting the accuracy and speed requirements.\par
	Most of the available vision-based traffic sensing systems process images to identify and count vehicles in each image captured. This process is complicated, time-consuming, and generally not implementable on general-purpose electronic hardware. On the other hand, there are well-established edge-detection methods that are not only fast but also implementable on general-purpose hardware. Therefore, this research attempts to use edge-detection methods on general-purpose electronic hardware to develop a new method that can estimate traffic intensity in real-time.
	\section{Vehicle Edge-Pixel Method for Incoming Traffic Intensity Estimation}
	\subsection{Rationale} 
\begin{figure}[htb]
		\includegraphics[width=0.8\linewidth]{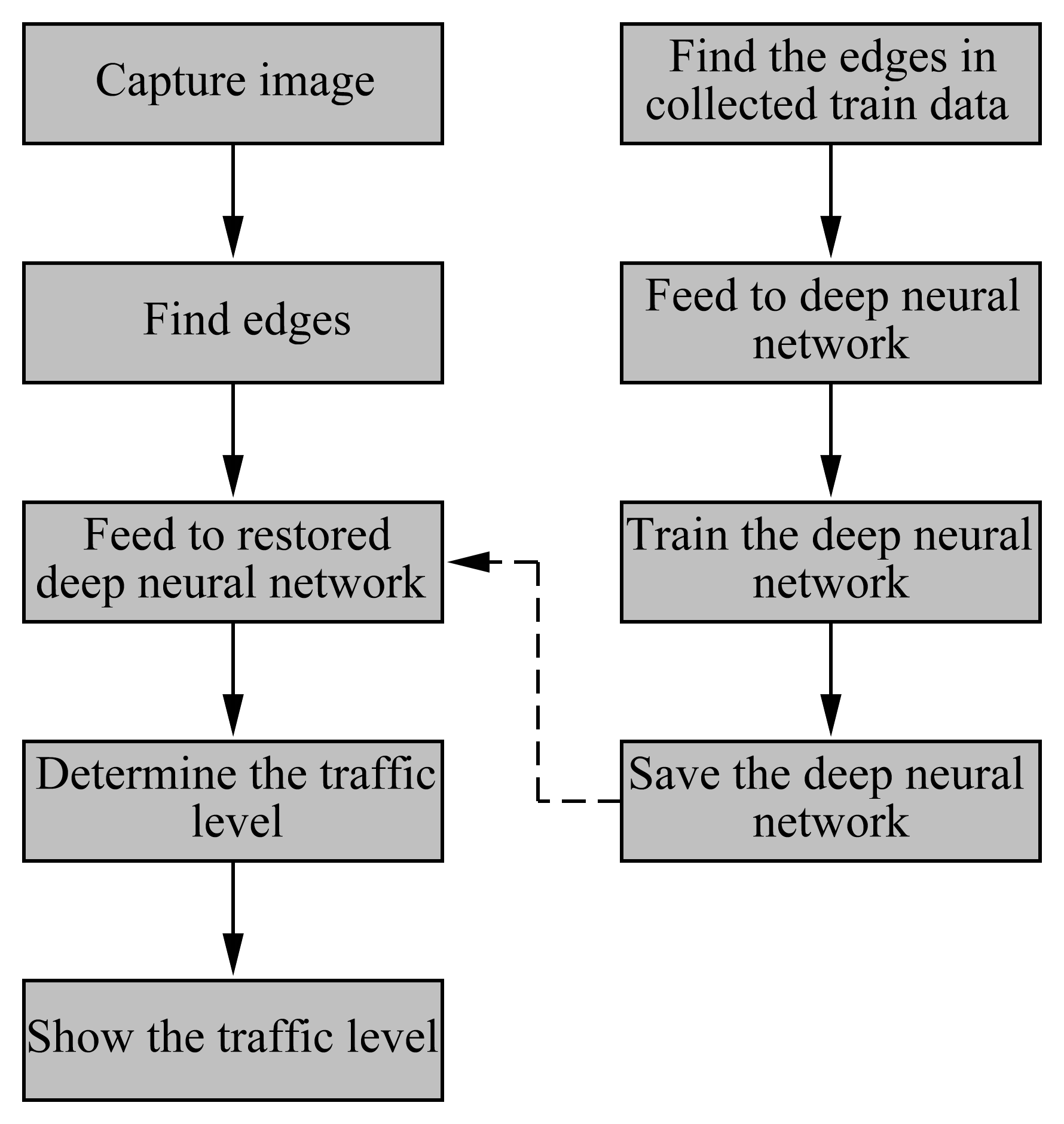}
		\centering
		\caption{Flow of the system.}
		\label{fig1}
	\end{figure}

A large vehicle creates a higher traffic intensity. A large vehicle also creates a higher edge-pixel count in the image. Hence, it is logical to hypothesize that there should be a way to estimate traffic intensity from the edge-pixel count in an image. And, this relationship would be nonlinear, and very difficult to model. It is also envisioned that the presence of vehicles in near, mid, and far sections of the road can be assessed with respective significance to estimate the incoming traffic intensity of a junction so that more accurate signal timing can be determined for the next signal cycle. This is in fact, feedforward traffic intensity estimation. In this context, a deep neural network is the most appropriate computational entity that can handle relevant input data and get trained to estimate the incoming traffic intensity. 
	Figure ~\ref{fig1} illustrates the edge detection-based method proposed in this research.
	The process has two parts. On the right is the offline supervised training of the deep neural network (DNN), and on the left is the real-time implementation of the trained DNN. The reason to use a neural network to determine the traffic intensity is based on the observation of a nonlinear relationship between edge count and the traffic intensity \cite{Neural_networks_in_manufacturing}. In the training process, a set of traffic images were taken using a Raspberry Pi camera. These images were visually inspected by an expert and assigned a number $\{$1,2,3,4,5$\}$ representing the traffic intensity in each image.
\subsection{Determination of Traffic Intensity Using Edge Pixels}
\begin{figure}[htb]
		\includegraphics[width=1.0\linewidth]{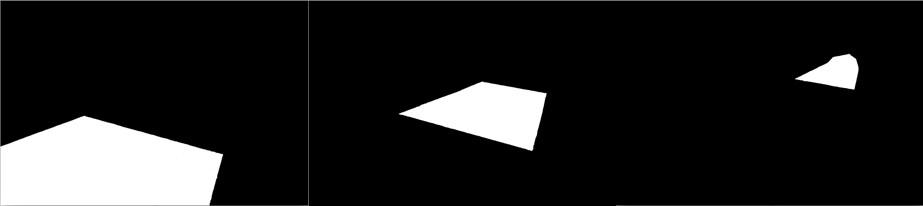}
		\centering
		\caption{Three zones on the road; near, mid, and far, from the intersection}
		\label{threezones}
	\end{figure}

\begin{figure}[htb]
		\includegraphics[width=1.0\linewidth]{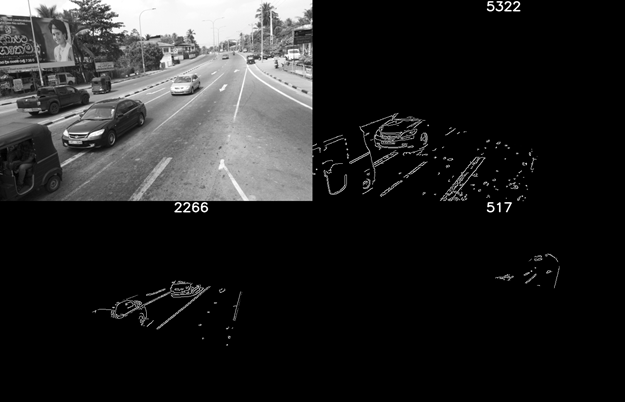}
		\centering
		\caption{Extraction of the road and the three zones for each image and detecting edges caused by vehicles}
		\label{photoandthreezones}
	\end{figure}

\begin{figure}[htb]
		\includegraphics[width=1.0\linewidth]{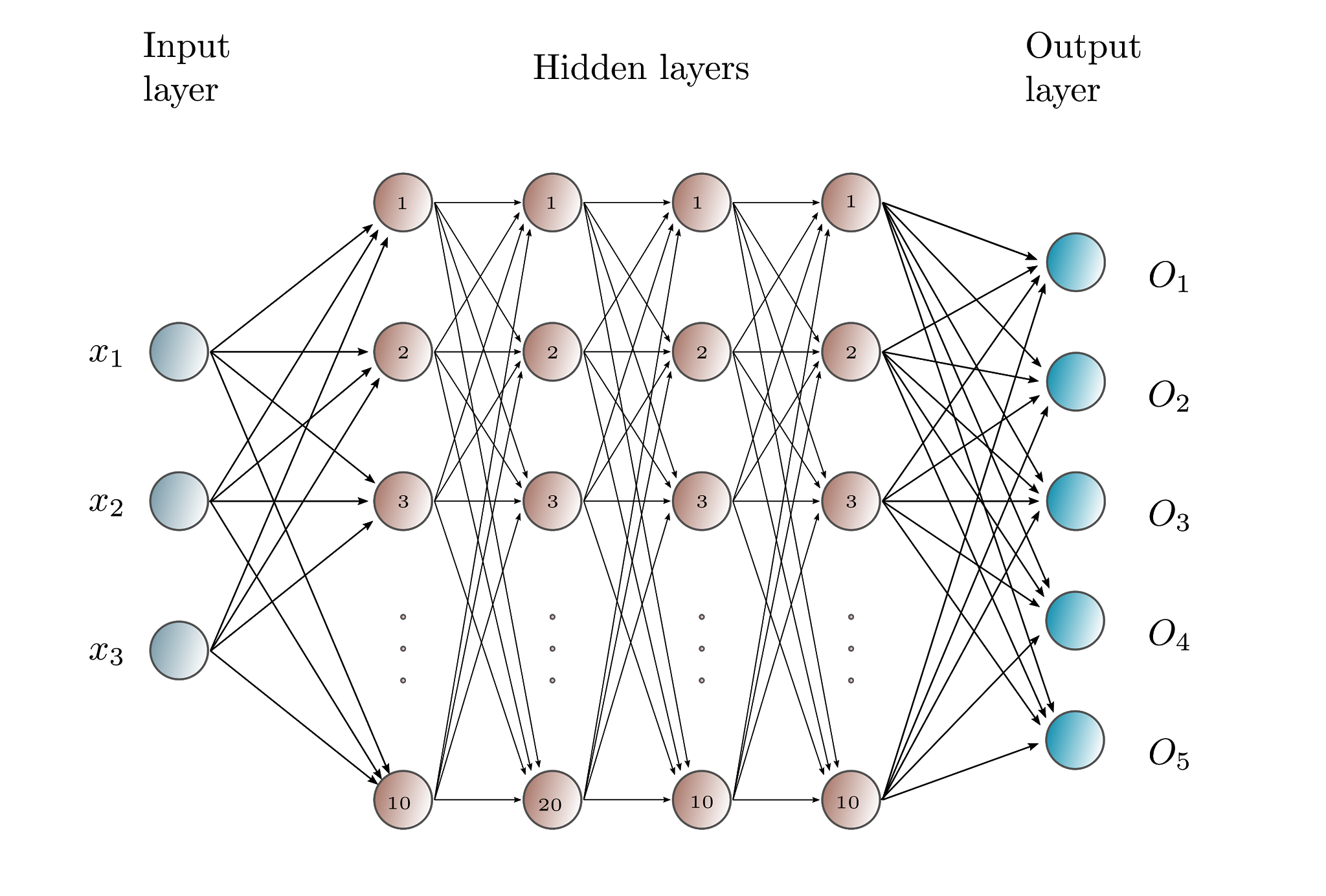}
		\centering
		\caption{Deep neural network (DNN) used for the determination of real-time traffic intensity}
		\label{dnn}
	\end{figure}

The field of view of the camera pointed along the road was divided into three zones; ``near'', ``mid'', and ``far''. The edges of the vehicles in each zone in the image are detected and the total number of pixels in those edges are counted. The three zones defined are shown in Fig.\ref{threezones}
	Fig.\ref{photoandthreezones} shows how the road and the three zones are extracted from an image before detecting the edges of the vehicles in the image.
	The number of edge pixels of the three zones (near, mid, and far) of an image is used as input data to the DNN. The DNN network shown in Fig.\ref{dnn} consists of three input neurons to receive these three pixel counts. The output of the DNN consists of five neurons. Each neuron represents the traffic intensity from one to five. The average of those five neurons is considered the traffic intensity. There are four hidden layers consisting of 10, 20, 10, and 10 neurons in each from input to output. The Tensorflow method \cite{Data_classification_with_deep_learning_using_Tensorflow} was used to implement the DNN.
	During the raining stage, a set of images was annotated by an expert on a scale of 1(very low) to 5(very high) traffic intensity.
	\subsection{Expert Judgement of Incoming Traffic Intensity} The Following rules were adopted by the expert in assigning a traffic intensity in the scale of $\{$1,2,3,4,5$\}$ by inspecting the presence of vehicles in the near, mid, and far regions.
	
\begin{figure}[htb]
		\includegraphics[width=0.4\linewidth]{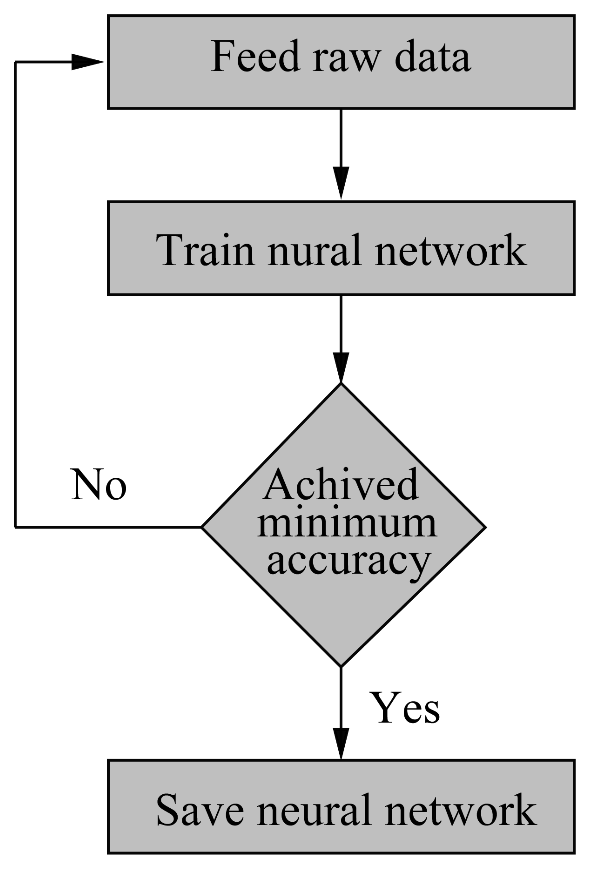}
		\centering
		\caption{Deep neural network training process.}
		\label{fig2}
	\end{figure}

	\begin{enumerate}
		\item The Presence of big vehicles particularly in the near and mid sections corresponds to high incoming traffic intensity.
		\item Small vehicles such as motorcycles and three-wheelers do not contribute as much traffic though they create a high edge-pixel density due to the vivid features they possess. Therefore, an edge-pixel count of more than 6000 in the near section, which is only possible with the presence of motorcycles and three-wheelers, alone may not indicate high traffic intensity.
		\item In the far section, the entire vehicle may be visible as a small image. This increases the edge-pixel density but with a lower total edge-pixel count. Therefore, the edge pixels in the far section have a higher contribution to the incoming traffic.
		\item When there are many vehicles in all three sections, traffic intensity is at level 5. When there are a few vehicles in the mid and far fields the traffic intensity is at level 1. The other levels 2, 3, and 4 are accordingly judged.
	\end{enumerate}
	The DNN was trained using a set of images that was annotated by the expert. An 80\% of the images were used to train the DNN and the rest 20\% was used for accuracy evaluation. Figure.~\ref{fig2} shows the training process, which is a recursive algorithm that loops until the error reduces to a given value. Training took place on a high-end computer.
	
	DNN training took place on a high-speed computer. The trained DNN was copied from the computer to the Raspberry Pi board, which was brought to the road intersection for real-time traffic intensity estimation.\par
\begin{figure}[htb]
		\includegraphics[width=0.55\linewidth]{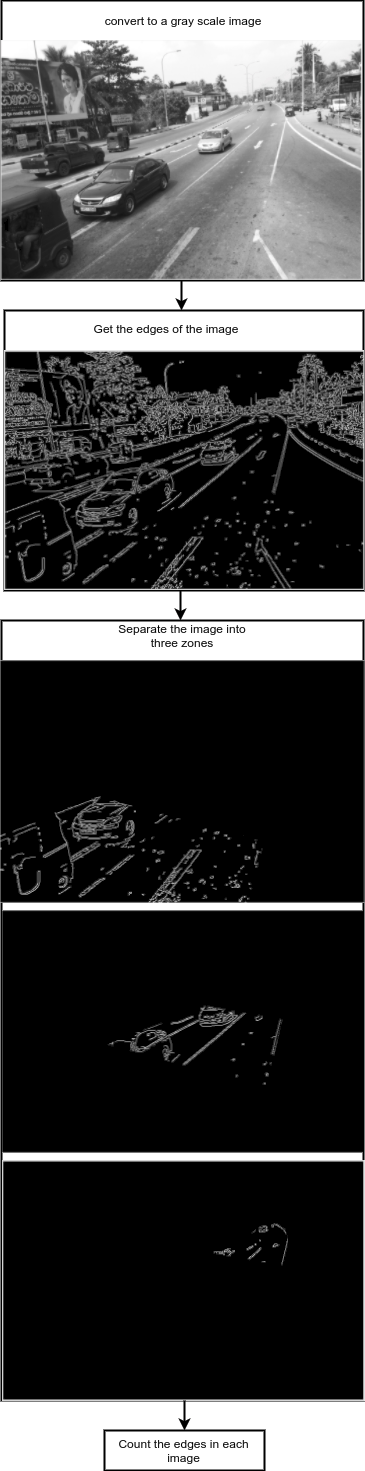}
		\centering
		\caption{Flow of the finding edges.}
		\label{fig3}
	\end{figure}

	Edge detection methods are widely used in image processing. These algorithms are not too complex, hence Raspberry Pi can run them efficiently while running the trained DNN. After comparing with other edge detection algorithms, the Canny edge detector was chosen after verifying the best performance consistently \cite{Edge_detection_revisited}, \cite{Comparative_analysis_of_common_edge_detection_techniques_in_context_of_object_extraction},\cite{Contextual_and_noncontextual_performance_evaluation_of_edge_detectors}. The image processing workflow from image capture to edge-pixel counts is shown in Fig.~\ref{fig3}, which involves the following steps.
	\begin{enumerate}
		\item Image is converted into grayscale
		\item Edges are detected using the Canny edge detector
		\item Image is cropped to near, mid, and far zones
		\item Count the edge pixels in each zone
	\end{enumerate}
	
	The process results in the edge pixel count in the near, mid, and far zones of the image. These three pixel counts along with the traffic intensity, as judged by an expert, are submitted for the supervised training of the DNN.
	\subsection{Removing Non-Vehicular Edges} 
\begin{figure}[htb]
		\includegraphics[width=1.0\linewidth]{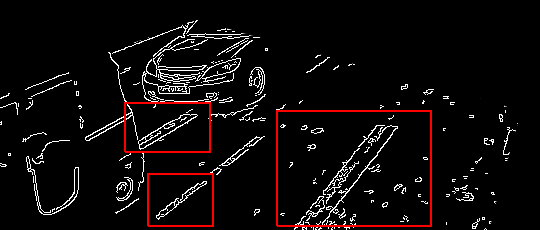}
		\centering
		\caption{Lane marks that are detected as edges}
		\label{fig4}
	\end{figure}

After initial training, it was realized that the neural network finds it difficult to recognize traffic level 1 (lowest) but always produces level 2. It was found that the features shown in Fig.~\ref{fig4} such as lane marks on the road that were erroneously detected as vehicle edges caused this issue. Such features usually are obscured by the vehicles when there are many vehicles in the near field (traffic levels 2 and above), but when there are just a few vehicles (traffic level 1) in the near field, these features appear and get counted to produce a higher-than-actual traffic intensity. Therefore, a new algorithm shown in Table.~\ref{table:1} was incorporated to remove these marks from the image.\par
	
\begin{figure}[ht]
		\includegraphics[width=1.0\linewidth]{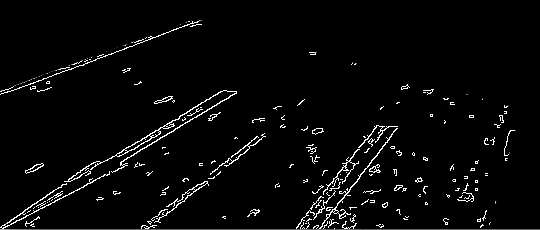}
		\centering
		\caption{The zero-traffic image without vehicles showing road marks and noise}
		\label{fig5}
	\end{figure}

	To remove the non-vehicular edges, the real-time image is compared with a zero-traffic image as shown in Fig.~\ref{fig5}. In this process, both images are compared pixel-to-pixel and evaluated as in Table.~\ref{table:1} where the pixels that appear black in zero-traffic images and turn white in the real-time image are counted as genuine edge pixels that are caused by vehicles.
	
\begin{figure}[htb]
		\includegraphics[width=1.0\linewidth]{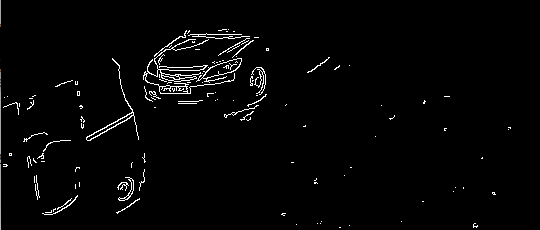}
		\centering
		\caption{An image after removing erroneous edges}
		\label{fig6}
	\end{figure}

	\begin{table}[htb]
	\centering
	\caption{Erroneous Edge Pixel Removing Criterion}
	\begin{tabular}{|p{2.5cm}|p{2.2cm}|p{2.0cm}|} \hline
	Pixel of the zero-traffic image & Corresonding pixel of the realtime image & Decision \\ [0.5ex] \hline
	black & black & Don't count \\ \hline
	black & white & Count \\\hline
	white & white & Don't count \\\hline
	white & black & Don't count \\\hline
	\end{tabular}
	\label{table:1}
	\end{table}
	The pixels that are white in the zero-traffic image are surely due to road marks and noise so they must not be counted. However, if a pixel that is black in the zero-traffic image becomes white in the real-time image it should be due to a vehicle edge. The erroneous edge-pixel removing algorithm is as follows:
	\begin{algorithm}
		\label{algo1}
		\caption{Removing Erroneous Edges}\label{euclid}
		\begin{algorithmic}
			\Require : input
			\Ensure : output
			\For{$position$ in $zero\_traffic\_image$}
			\State $height$ = $position$[0]
			\State $width$ = $position$[1]
			\If {($zero\_traffic\_image[h,w]==0$ \& $real\_time\_image[h,w]==255$)}
			\State create a $kernel$ in $zero\_traffic\_image$
			\If {(at least one pixel is white in $kernel$)}
			\State $edge\_count += 1$;
			\EndIf
			\EndIf
			\EndFor
		\end{algorithmic}
	\end{algorithm}
	Figure~\ref{fig6} shows the real-time image after removing the no-nvehicular edge-pixels  It is quite obvious that the genuine vehicular edge-pixels are intact while almost all of the non-vehicular edge-pixels have disappeared.
	\section{Results}
	\subsection{DNN Accuracy in Estimating Traffic Intensity}
	The accuracy of the DNN in estimating traffic intensity was determined by comparing the DNN estimate against the expert's judgment using a set of over 300 images, in which 64\% was correctly estimated by the DNN while in another 34\%, the DNN estimate was one level different to that of the expert’s.
	
	
	\begin{table}[htb]
		\centering
		\caption{Results}
		\begin{tabular}{p{2.2cm}p{2.2cm}p{2.2cm}} \hline
		\vspace{1mm}\centering Image & \vspace{1mm}\centering Expert's Judgement & \vspace{1mm} DNN Estimate \\ 
		\includegraphics[width=1.0\linewidth]{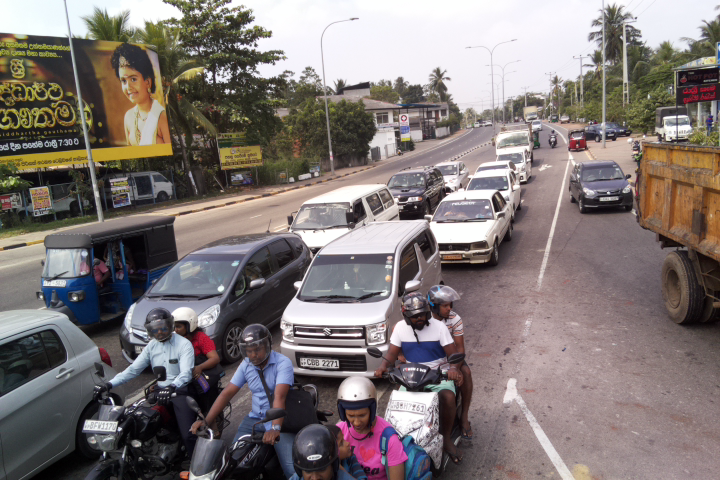} & \hspace{1cm}5 & \hspace{1cm}3.7 \\ 
		\includegraphics[width=1.0\linewidth]{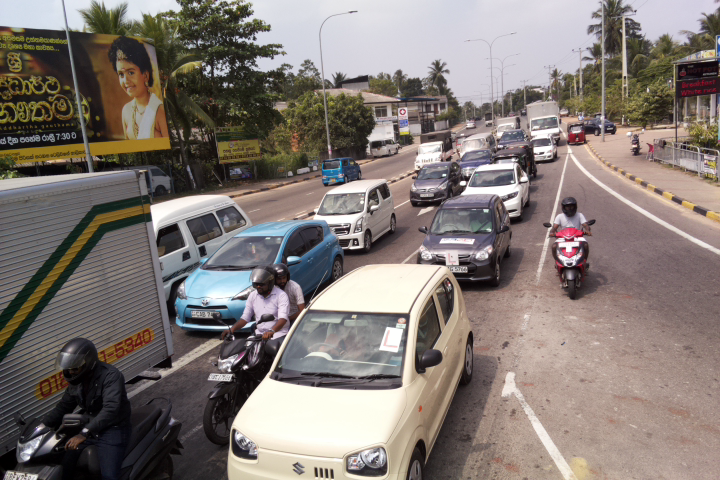} & \hspace{1cm}5 & \hspace{1cm}3.9 \\ 
		\includegraphics[width=1.0\linewidth]{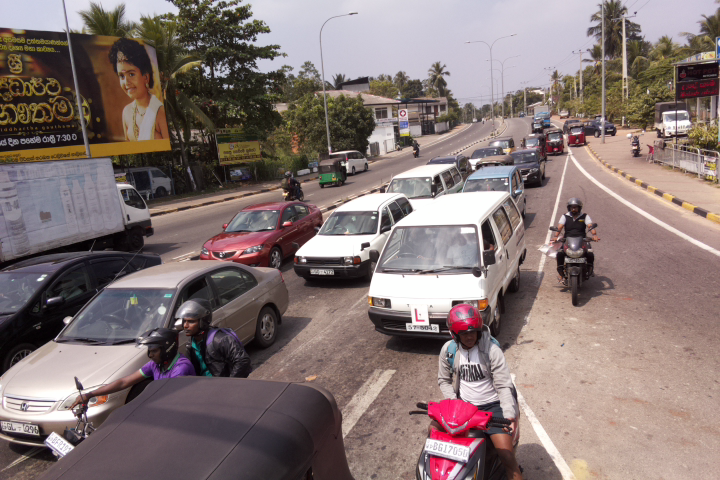} & \hspace{1cm}4 & \hspace{1cm}3.6 \\
		\includegraphics[width=1.0\linewidth]{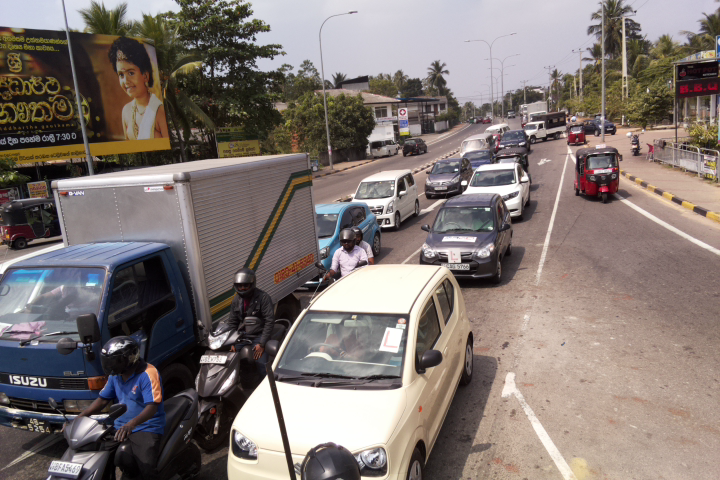} & \hspace{1cm}4 & \hspace{1cm}3.9 \\
		\includegraphics[width=1.0\linewidth]{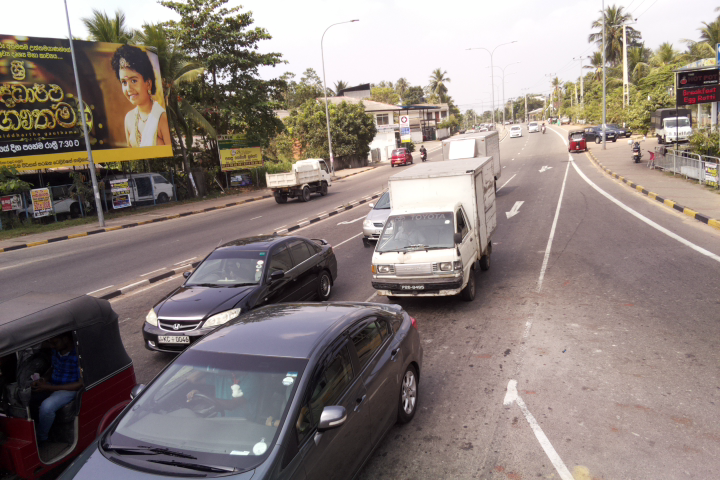} & \hspace{1cm}3 & \hspace{1cm}2.9 \\
		\includegraphics[width=1.0\linewidth]{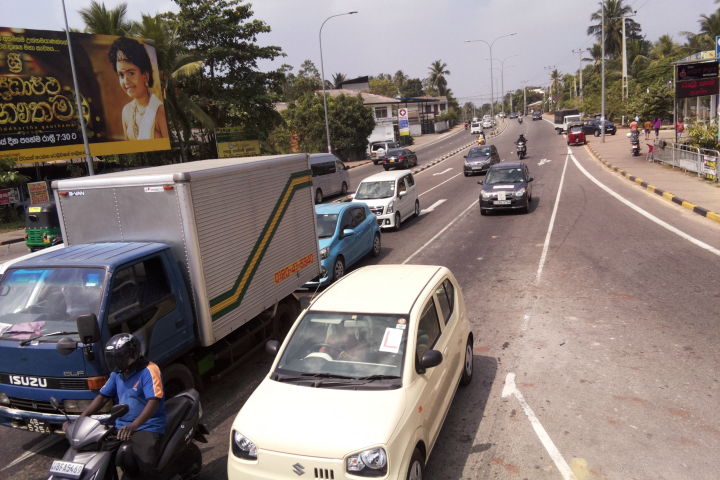} & \hspace{1cm}3 & \hspace{1cm}2.0 \\
		\includegraphics[width=1.0\linewidth]{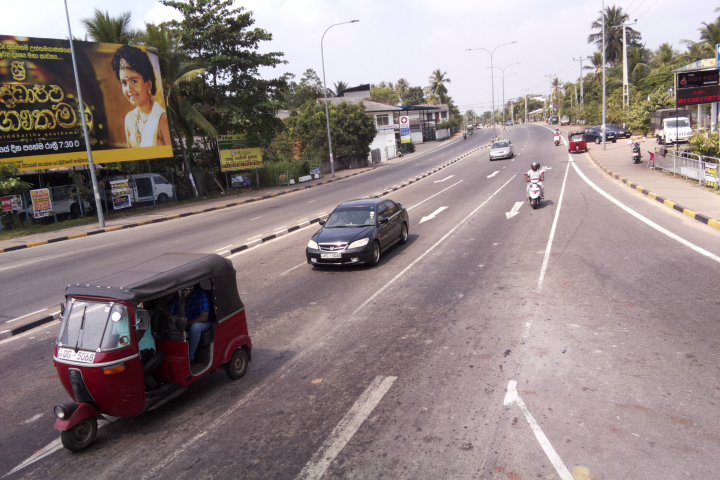} & \hspace{1cm}2 & \hspace{1cm}1.9 \\
		\includegraphics[width=1.0\linewidth]{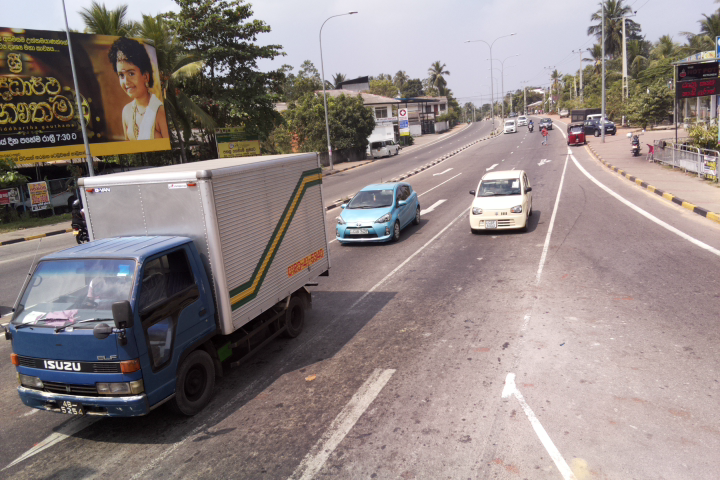} & \hspace{1cm}2 & \hspace{1cm}2.4 \\
		\includegraphics[width=1.0\linewidth]{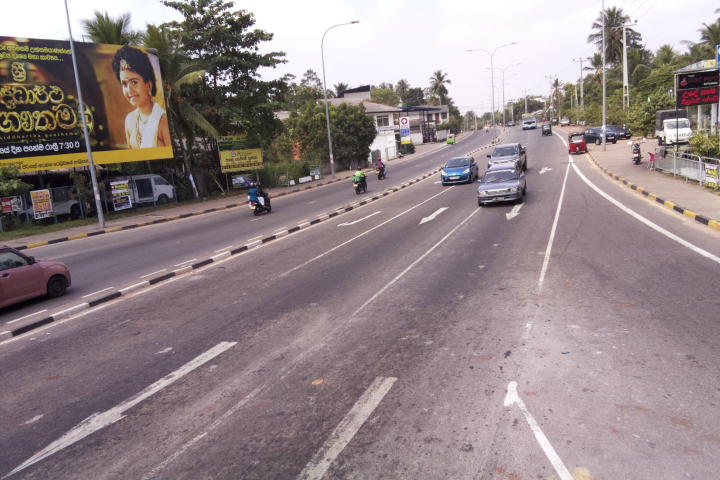} & \hspace{1cm}1 & \hspace{1cm}1.0 \\
		\includegraphics[width=1.0\linewidth]{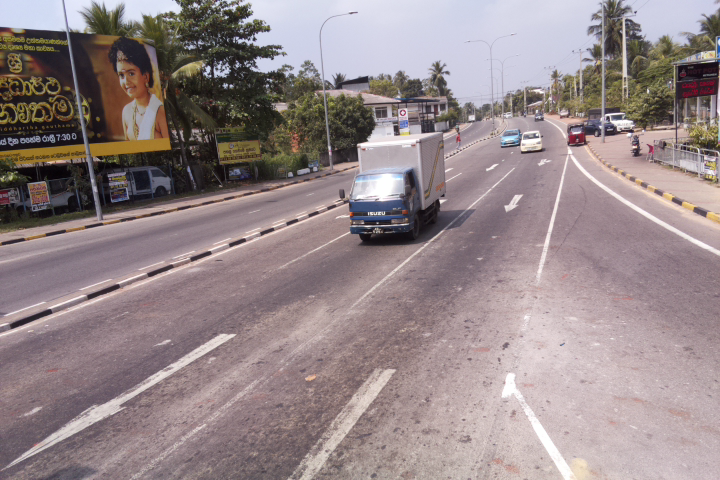} & \hspace{1cm}1 & \hspace{1cm}1.0 \\ \hline 
		\end{tabular}
		\centering
		\label{table:2}
	\end{table}
	\begin{table}[htb]
		\centering
		\caption{Rate of accurate Estimate of Traffic Intensity by the DNN}
		\begin{tabular}{|p{2.2cm}|p{1.2cm}|} 
			\hline
			Perfect Judgement  & 64\% \\ [0.5ex] 
			\hline
			One step difference  & 30\% \\ 
			\hline
		\end{tabular}
		\centering
		\label{table:3}
	\end{table}
	\subsection{Processing Time}
	Traffic intensity estimation has the following steps: (1) image capture, (2) Edge detection, (3) removing false edge pixels and DNN process, and (4) Display traffic intensity. In the first attempt with standard coding, it took about 20s to process from image capture to the determination of traffic intensity. Then, the image processing was implemented parallelly to speed up the process. Table \ref{table_speed} shows the time durations after introducing parallel processing.
	\begin{table}[htb]
	\centering
	\caption{Processing Speed}
	\begin{tabular}{|p{1.9cm}|p{0.6cm}|p{0.6cm}|p{0.6cm}|p{0.6cm}|p{0.6cm}|p{0.6cm}|} \hline
	Process & Im1(s) & Im2(s) & Im3(s) & Im4(s) & Im5(s) & Im6(s)\\ [0.5ex] \hline
	Image capture & 1.1 & 1.1 & 1.1 & 1.0 & 1.1 & 1.1\\ \hline
	Edge-detection & 1.4 & 3.2 & 2.6 & 1.7 & 2.1 & 1.7\\ \hline
	DNN Processing & 0.1 & 0.2 & 0.2 & 0.2 & 0.2  & 0.2\\ \hline
	Display Result & 2.0 & 2.0 & 2.0 & 2.0 & 2.0 & 2.0\\ \hline
	Total & 4.7 & 6.5 & 5.9 & 4.9 & 5.4 & 5.0\\ \hline
	\end{tabular}
	\centering
	\label{table_speed}
	\end{table}
	Neglecting the display time of 2s in each image, the average time from image capture to the determination of the traffic intensity is 3.4s. This is fast enough to be implemented in real-time adaptive traffic control.
	\subsection{Real-time Testing on the Road}
\begin{figure}[htb]
		\includegraphics[width=1.0\linewidth]{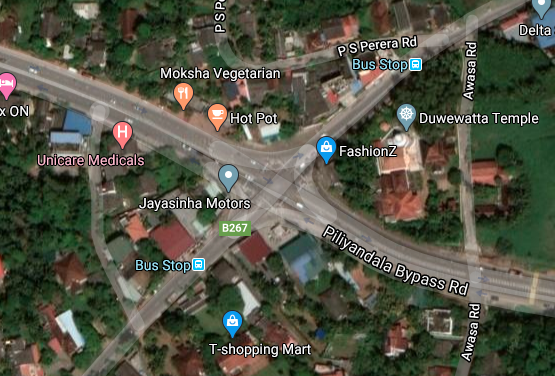}
		\centering
		\caption{The junction where real-time testing was conducted. Location 79$^{\circ}$56’6.05”E and 6$^{\circ}$48’4.70”N}
		\label{junction}
	\end{figure}

	The trained DNN on a Raspberry Pi single-board computer was placed on a traffic signal post at the three-lane four-way junction shown in Fig.\ref{junction} at Piliyandala outer circular road.
	
\begin{figure}[htb]
		\includegraphics[width=1.0\linewidth]{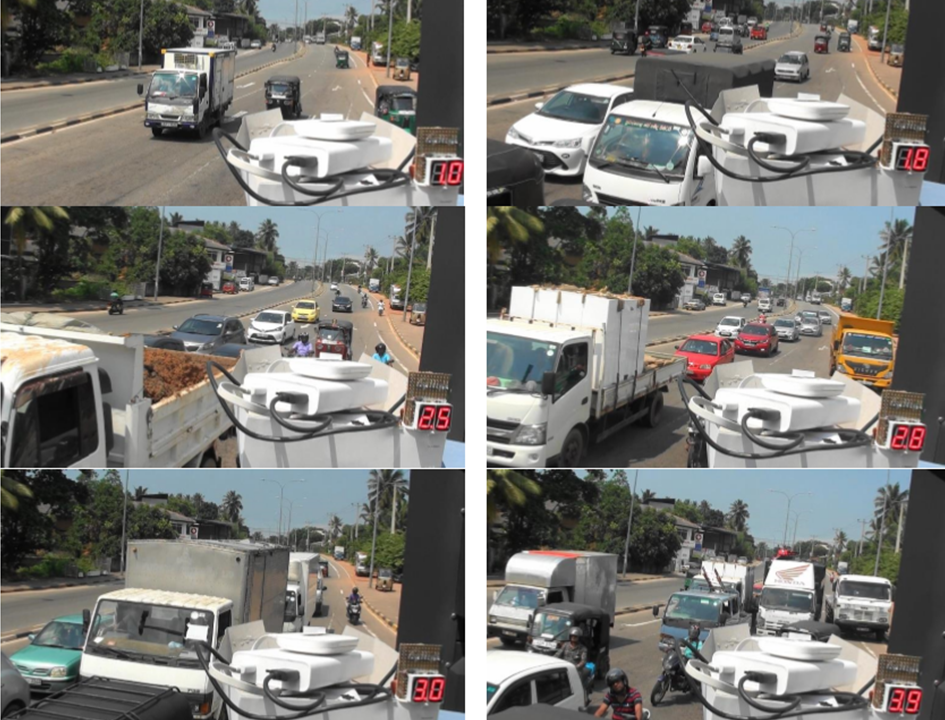}
		\centering
		\caption{Results of the road test}
		\label{roadtest}
	\end{figure}

	Raspberry Pi was connected to a seven-segment display to indicate incoming traffic intensity on the scale of [1,2,3,4,5]. The test went on for many sessions covering morning and evening rush hours and daytime moderate traffic conditions. Figure.\ref{roadtest} shows the real-time results of this test.

	\section{Conclusion}
	In this research, a novel vision-based real-time traffic intensity estimation method was developed using machine learning on general-purpose embedded electronic hardware. The method uses edge detection instead of vehicle identification, and using the vehicular edge-pixel counts in the near, mid, and far sections of the road the incoming traffic was estimated using a trained deep neural network, which showed 64\% accuracy in perfect estimation in the traffic intensity in the range of [1,2,3,4,5]. In another 34\% of the images, the DNN estimate was just one level different from the expert's judgment. Considering challenges in defining traffic intensity precisely, and the uncertainty of traffic intensity when it falls between adjacent levels, the 34\% of instances where the DNN estimate was just one level off could still be used for adaptive traffic control. The mean processing time was 3.4s, hence the method can be used for real-time traffic control.\par
	The ability to use general-purpose electronic hardware makes the method cost-effective and financially viable for mass manufacturing and deployment even in suburban areas. 
	
	\section{Acknowledgement}
	This work was initially funded by the Innovation Quotient (Pvt) Ltd., and later through the World Bank's AHEAD project. Prof. Saman Bandara of the Traffic Engineering Division of the University of Moratuwa provided valuable insights while Mr. Chiranthaka Kapukotuwe contributed with initial data gathering.

\balance
\end{document}